%% file: main.tex
\RequirePackage[svgnames,table]{xcolor}

\documentclass[11pt,letterpaper,copyright]{style1}

\renewcommand{\copyrightext}{%
  \copyrightfont
  \textcopyright{} 2026 JP Morgan Chase \& Co. All rights reserved.%
}

\input{preamble}   
\input{command}    

\definecolor{LPMAccent}{HTML}{8F5A39}
\definecolor{LPMTint}{HTML}{F8F4F1}
\definecolor{LPMTable}{HTML}{FEF4DB}

\renewcommand{\titlefont}{\centering\color{LPMAccent}\normalfont\bfseries\fontsize{18}{20}\selectfont}
\newcommand{\lpmemphasis}[1]{\textcolor{LPMAccent}{\textbf{#1}}}
\newcommand{\lpmindependent}{\texttt{Independent}}
\newcommand{\lpmsequence}{\texttt{Sequence}}

\tcbset{
    lpmCallout/.style={
        enhanced,
        breakable,
        colback=White,
        frame hidden,
        borderline west={2pt}{0pt}{LPMAccent},
        sharp corners,
        boxsep=0pt,
        left=10pt, right=0pt,
        top=2pt, bottom=2pt,
        before skip=10pt,
        after skip=10pt,
    }
}
\newtcolorbox{lpmDefinition}{lpmCallout}
\newtcolorbox{lpmTakeaway}[1][]{lpmCallout,#1}

\tcbset{
    titlebox/.style={
        colback=LPMTint,
        colframe=LPMAccent!8!White,
        boxrule=0.5mm,
        arc=2mm,
        auto outer arc,
        left=5mm,
        right=5mm,
        top=5mm,
        bottom=5mm,
    }
}

\newtcolorbox{simpleElegantQuote}{
    colback=LPMTint,
    colframe=LPMAccent,
    boxrule=0.5pt,
    arc=2mm,
    boxsep=4pt,
    left=10pt, right=10pt,
    top=8pt, bottom=8pt,
    fontupper=\itshape,
}

\title{Large Payments Model}

\runningtitle{Large Payments Model - JPMorgan Chase \& Co}


\usepackage{fontawesome5}   

\newcommand{\lpmauthorrow}[1]{%
    \parbox[c][\baselineskip][c]{\linewidth}{%
        \makebox[\linewidth]{\hspace{0em}#1\hspace{0em}}%
    }\par
}

\author{%
\mdseries
\setlength{\parskip}{0pt}%
\lpmauthorrow{Natraj Raman\textsuperscript{*}\hfill Abhenandhan Manoharan\hfill Vishwasreddy Thuniki\hfill Dhvani Kansara}
\lpmauthorrow{Anirudh Som\hfill Hassan Firouzi\hfill Kostis Gourgoulias\hfill Laukik Hoshing\hfill Sudhir Kanala\hfill Austin Reiter}
\lpmauthorrow{Ebby Abdul\hfill Shantanu Chandra\hfill Ryan Schmiedl\hfill Zack Anderson\hfill Sumitra Ganesh}
\raisebox{-4pt}{\makebox[\linewidth][c]{\textbf{JPMorgan Chase \& Co}}}\\
}

\hypersetup{
    pdftitle={Large Payments Model},
    pdfauthor={Natraj Raman},
    pdfsubject={A domain-specific foundation model for payment transactions},
    pdfkeywords={domain foundation models, payment transactions, structured data,
        fraud detection},
    pdfdisplaydoctitle=true,
    colorlinks=true,
    linkcolor=blue!50!black,
    citecolor=blue!50!black,
    urlcolor=blue!50!black
}

\begin{document}

\input {sec_abstract}

\maketitle

\iftrue
\begingroup
\renewcommand{\thefootnote}{\fnsymbol{footnote}}
\footnotetext[1]{Corresponding author: \texttt{natraj.raman@jpmorgan.com}}
\endgroup
\fi

\input {sec_intro}

\input {sec_relwork}
\input {sec_model}

\input {sec_adap}

\input {sec_experiments}
\input {sec_deploy}
\input {sec_findings}
\input {sec_conclusion}

\section*{Acknowledgments}
This paper was prepared for informational purposes by the Artificial Intelligence Research group of JPMorgan Chase \& Co. and its affiliates ("JP Morgan") and is not a product of the Research Department of JP Morgan. JP Morgan makes no representation and warranty whatsoever and disclaims all liability, for the completeness, accuracy or reliability of the information contained herein. This document is not intended as investment research or investment advice, or a recommendation, offer or solicitation for the purchase or sale of any security, financial instrument, financial product or service, or to be used in any way for evaluating the merits of participating in any transaction, and shall not constitute a solicitation under any jurisdiction or to any person, if such solicitation under such jurisdiction or to such person would be unlawful. © 2026 JP Morgan Chase \& Co. All rights reserved.

\bibliography{references}
\bibliographystyle{plainnat}


\end{document}

%% file: preamble.tex
\usepackage{graphicx}
\usepackage{float,epstopdf}
\usepackage{bbm}

\usepackage{microtype}

\usepackage{natbib}
\setcitestyle{square}

\usepackage{subcaption}
\usepackage{booktabs}

\usepackage{amsmath}

\usepackage{amssymb}
\usepackage{mathtools}
\usepackage{amsthm}
\usepackage{dsfont}
\usepackage{multicol}
\usepackage{makecell}
\usepackage{multirow} 
\usepackage{amsfonts} 
\usepackage{mathrsfs}
\usepackage[amssymb, thickqspace]{SIunits}
\usepackage{enumitem}
\usepackage{pgfplotstable}
\usepackage{lipsum}		

\usepackage{microtype}
\usepackage{graphicx}
\usepackage{booktabs} 
\usepackage[table]{xcolor}
\usepackage{arydshln}
\usepackage[normalem]{ulem} 

\usepackage{cases}
\usepackage{wrapfig}

\usepackage{url}

\usepackage{thmtools}
\usepackage{thm-restate}

\definecolor{huskypurple}{HTML}{4B2E83}

\usepackage{titletoc}

\usepackage{listings}
\lstdefinestyle{promptstyle}{
  basicstyle=\ttfamily\footnotesize,
  breaklines=true,
  breakautoindent=false,
  breakindent=0pt,
  postbreak=\mbox{\textcolor{gray}{$\hookrightarrow$}\space},
  columns=fullflexible,
  keepspaces=true,
  frame=single,
  framesep=5pt,
  xleftmargin=6pt,
  xrightmargin=6pt,
  aboveskip=8pt,
  belowskip=8pt,
  showstringspaces=false,
}

%% file: command.tex
\makeatletter
\def\munderbar#1{\underline{\sbox\tw@{$#1$}\dp\tw@\z@\box\tw@}}
\makeatother

\defcitealias{stripe_pfm}{PFM}
\defcitealias{lloyds_atum}{Atum}

\AddToHook{cmd/appendix/before}{%
  \setcounter{axiom}{0}%
}

\newcommand{\be}{\begin{equation}}
\newcommand{\ee}{\end{equation}}
\newcommand{\bea}{\begin{equation*}\begin{aligned}}
\newcommand{\eea}{\end{aligned}\end{equation*}}



%% file: sec_abstract.tex
\begin{abstract}
\vspace{-1mm}
{\centering\section*{Abstract}}
Payment applications such as fraud detection, authorization prediction, 
and payment optimization draw on overlapping transaction information. 
This motivates a reusable payments intelligence layer that shares learned 
representations and predictive signals, rather than developing predictive models 
independently for each task. Realizing this shared capability requires handling
heterogeneous transaction fields, high-cardinality identifiers,
lifecycle-dependent attributes, long histories, and rare outcomes while
meeting latency and cost constraints. We present
\emph{Large Payments Model (LPM)}, a
domain-specific foundation model designed for operational use and trained from
scratch on large-scale structured transaction data to learn reusable
representations for payment prediction tasks. LPM employs field-specific
encodings and lifecycle-aware self-supervision in two complementary
architectures: the
\lpmindependent{} variant uses hybrid masking and supports efficient
per-transaction inference, while the \lpmsequence{} variant incorporates
entity-linked histories through compact latent representations for richer
temporal context. Optional distillation enables resource-constrained
deployment. We pretrain LPM on up to 400 billion tokens and study scaling,
downstream performance, and representation reuse. LPM improves over strong task-specific
baselines across multiple prediction tasks in Merchant and Treasury Services,
demonstrating applicability in distinct payment domains.
Its representations also facilitate adaptation to a target not observed
during pretraining through compact task-specific components. Together with
production deployment, these results position LPM as a practical basis for
reusable payments intelligence.
\end{abstract}

%% file: sec_intro.tex
\section{Introduction}
\label{sec:intro}

Payment activity produces rich records spanning authorization, clearing,
settlement, returns, and disputes. These records provide a shared source of information for fraud detection
~\citep{hafez2025systematic}, authorization prediction, retry assessment, returns analysis, and
payment optimization. Developing
predictive models independently for these capabilities can require relearning
overlapping transaction structure and behavior. A reusable payments
intelligence layer instead enables shared learning from payment activity,
producing representations and predictive signals that support multiple tasks.

Foundation models provide a precedent for this shift from task-specific
learning to reusable representations through large-scale pretraining on
predominantly unlabeled data. This paradigm has proved powerful for language, vision, speech, and code
 modalities ~\citep{zhou2025comprehensive, lu2025representation}. For payments, the
opportunity is to reuse learned knowledge of transaction structure and behavior
across prediction tasks through common pretraining, representation interfaces,
and serving infrastructure.

Realizing this opportunity requires treating payment records as a distinct
modality. Each transaction is a structured event composed of heterogeneous
fields---categorical identifiers, continuous amounts, timestamps, status
codes, and response codes---whose meaning is defined by their role in a fixed
schema. Transactions are also inseparable from the operational lifecycle and
temporal context in which they occur: an event may only be interpretable
relative to prior activity, and the information available at authorization
differs from what is observed later in the payment flow. While serializing
transactions as text can make them syntactically consumable by language models
~\citep{hegselmann2023tabllm}, it does not, by itself, encode field semantics,
lifecycle constraints, or temporal relationships across events. A payments
foundation model should therefore learn directly from the native structure of
transaction data rather than treating that structure as incidental text.

\begin{figure*}[t]
	\centering
	\includegraphics[width=\textwidth]{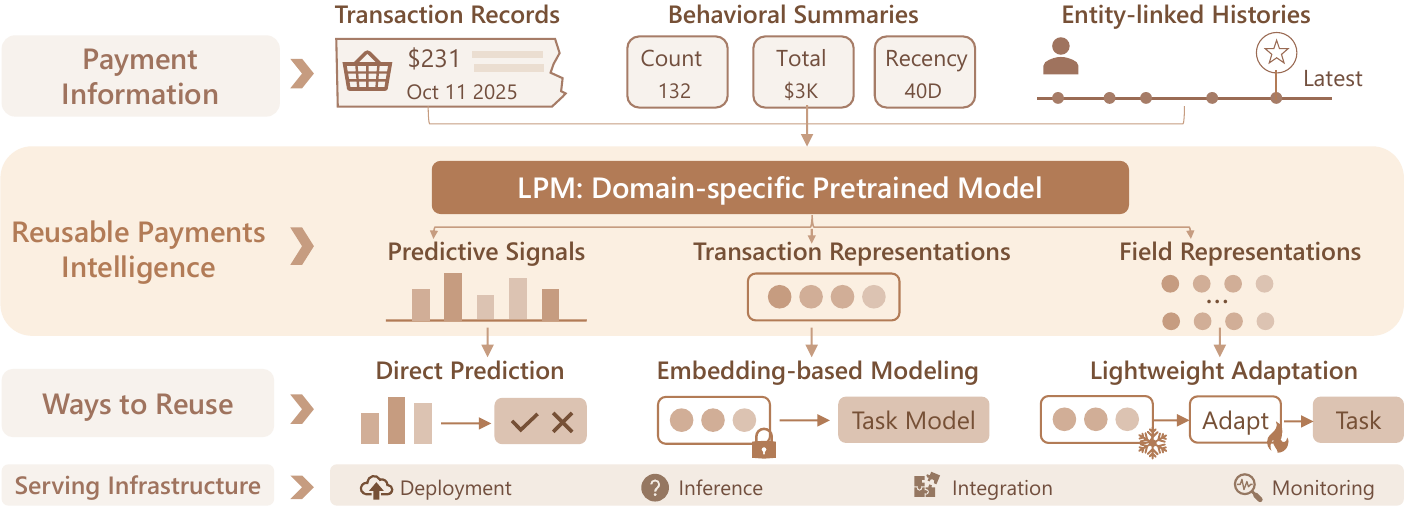}
	\caption{\textbf{From shared learning to downstream use.} A pretrained LPM provides a common starting point for payment tasks, reducing the need to learn transaction structure independently for each application. The reuse routes differ in the task-specific modeling they require, while shared serving infrastructure enables operational delivery.}
	\label{fig:lpm-reuse}
\end{figure*}

\begin{figure*}[t]
	\centering
	\includegraphics[width=\textwidth]{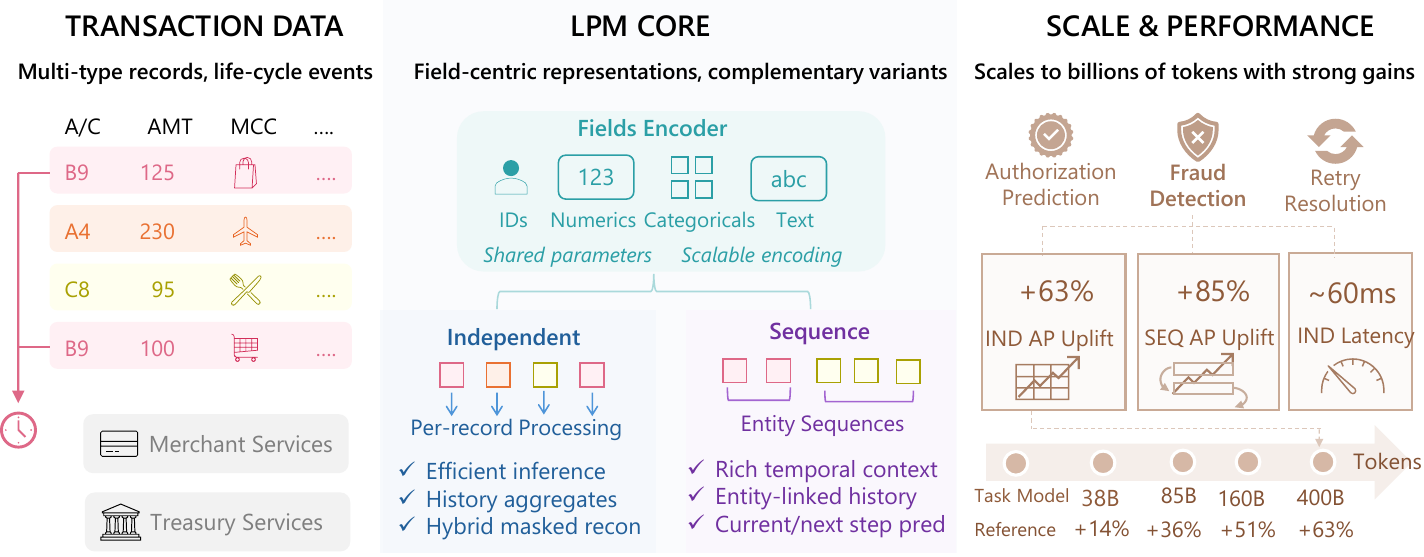}
	\caption{\textbf{LPM at a glance}. LPM encodes heterogeneous transaction fields with a field-aware encoder and supports two variants: an \lpmindependent{} model using aggregate context and hybrid masking, and a \lpmsequence{} model capturing entity-linked histories for temporal context. We evaluate LPM on Merchant Services and Treasury Services, highlighting relative improvements in Merchant Services fraud detection average precision over a task-specific baseline.}
	\label{fig:lpm-overview}
\end{figure*}

Building foundation models for structured data is challenging
~\citep{hollmann2025accurate}, and payment transactions introduce additional
domain-specific requirements. A model must represent high-cardinality
entities, such as cards and merchants, without allowing embedding tables to
dominate its size. It must capture both
within-transaction interactions and patterns across long histories at a
practical computational cost. Pretraining must also learn from lifecycle
outcomes without exposing information unavailable at the time of a downstream
decision. Finally, the resulting representations must transfer across tasks
and data scales while satisfying the latency, memory, and computational
constraints of payment systems.

\begin{lpmDefinition}
In this work, we present the \emph{Large Payments Model (LPM)}, a
domain-specific foundation model trained from scratch on large-scale structured
payment transaction data using self-supervised reconstruction. LPM learns
representations of transactions and their attributes that provide predictive
signals for a range of downstream payment tasks. Its serving architecture is
designed for low-latency, cost-efficient production inference. These
capabilities connect shared learning with downstream use and operational
delivery, positioning LPM as a reusable payments intelligence layer
(\autoref{fig:lpm-reuse}).
\end{lpmDefinition}

\lpmemphasis{Our design} addresses payments-specific requirements through complementary choices in representation, modeling, training, and deployment. Field-aware representations preserve the semantics of heterogeneous inputs while sharing parameters across fields where appropriate. Specialized numerical encodings retain fine-grained value information, while a non-uniform partitioned embedding mechanism ~\citep{raman2024scalable} allocates capacity selectively across categorical values instead of maintaining uniformly large embedding tables. Two model variants combine different temporal and pretraining strategies to balance contextual richness with inference cost. Optional distillation to compact students ~\citep{mansourian2025comprehensive} provides further flexibility under tighter deployment constraints.

The \lpmemphasis{\lpmindependent{} variant} jointly models fields within a
transaction while accepting historical context through fixed-size aggregate
features. Its inference cost is independent of raw history length, making the variant well suited to latency-sensitive applications. It is pretrained using a hybrid masking strategy ~\citep{tay2022ul2} that
combines random field masks, masks over semantically related groups, and
lifecycle-aware masks aligned with the information available at an operational
decision point.

The \lpmemphasis{\lpmsequence{} variant} directly models within-transaction interactions and temporal dependencies across transaction histories. It uses compact latent representations ~\citep{jaegle2021perceiver} and, inspired by the interleaved attention pattern of Latte
~\citep{ma2024latte}, alternates intra- and inter-transaction transformer
blocks. This structure incorporates historical context efficiently while retaining access to detailed transaction information. The variant is intended
for settings where the value of richer temporal context
justifies the additional data-access and computational requirements. \autoref{fig:lpm-overview}
summarizes the LPM framework, its two model variants, and the evaluation at scale.

\lpmemphasis{We evaluate} the breadth and reuse of LPM's predictive
representations. In Merchant Services, prediction heads learned during
pretraining improve fraud detection, Network Response, and Retry Response over
task-specific baselines, demonstrating utility across several payment
objectives. The \lpmindependent{} model improves fraud detection average
precision by about 60\%, and the \lpmsequence{} variant adds more than 20
percentage points of uplift when transaction history is available. Fine-grained Network
Response macro-F1 improves by 80\% or more. To assess applicability in a
distinct payment domain, we also evaluate a separately trained Treasury
Services model on ACH and wire fraud prevention, achieving about 27\% uplift
in ACH volume-based recall. 

\lpmemphasis{Beyond the pretraining targets}, we evaluate reuse of an existing
LPM backbone for eventual retry-approval prediction. A linear probe on frozen
transaction representations comes within about 2\% of a dedicated task-specific
baseline in average precision, while low-rank adaptation achieves comparable
performance with only 670K trainable parameters. This result
demonstrates extension to a target not observed during pretraining through
compact task-specific components. Controlled studies further characterize
representation structure, scaling, sequence length, and distillation. LPM also
operates at production scale, serving millions of transactions per day while
completing 99\% of requests within $60\,\mathrm{ms}$.

Together, task breadth, domain applicability, and new-target transfer provide
complementary evidence for reusable payments intelligence, while production
serving demonstrates the feasibility of delivering predictive signals at
payment scale.

%% file: sec_relwork.tex
\section{Related Work}
\label{sec:relwork}
Foundation models for structured data ~\citep{raman2026toward} have followed several directions.
LLMs can operate on serialized tables, but their performance is sensitive to
how structure is represented ~\citep{hegselmann2023tabllm,sui2024table,
liu2024rethinking}. General-purpose tabular foundation models pursue transfer
across tables or in-context prediction from labeled examples
~\citep{zhu2023xtab,hollmann2025accurate,grinsztajn2025tabpfn,
qu2026tabiclv2}. These approaches are effective for broad tabular learning,
but their generic representations and inference formulations do not directly
address payment-specific field semantics, delayed labels, or low-latency
event scoring. Time-series foundation models provide a complementary basis for
learning from histories, although much of this work emphasizes forecasting
numerical signals rather than representing heterogeneous, lifecycle-dependent
events ~\citep{kottapalli2025foundation,liang2024foundation}. Graph foundation
models offer another relevant direction, including few-shot anomaly
detection with potential applications to fraud ~\citep{qiao2025anomalygfm}.
At massive pretraining scale, however, irregular graph topologies can
complicate batching, partitioning, and distributed execution; message-passing
designs also face depth-related effects such as over-smoothing ~\citep{liu2025graph}.

Recent work---including TransactionGPT~\citep{dou2025transactiongpt}, nuFormer~\citep{braithwaite2025your}, TREASURE~\citep{yeh2026treasure}, PRAGMA~\citep{ostroukhov2026pragma}, and industrial initiatives~\citepalias{stripe_pfm,lloyds_atum}---treats financial transactions as a distinct modality and is closely aligned with our setting. LPM reflects a parallel domain-specific direction, focusing on native structured payment records, lifecycle-aware pretraining, and the operational trade-off between low-latency per-transaction inference and richer history-aware modeling. These priorities complement broader work on language, tabular,
time-series, and graph foundation models while addressing the representation and serving
requirements of payment systems.

%% file: sec_model.tex
\section{Model Architecture and Pretraining}
\label{sec:model}

LPM operates on payment records in their native structured form and learns
transaction representations that can be reused across prediction tasks. Its
architecture separates two questions: how heterogeneous fields are encoded,
and how much transaction context is available at inference. A shared
field-aware encoder maps schema-defined values to a common $D$-dimensional
representation space. The \lpmindependent{} model jointly contextualizes these field
representations within one transaction for efficient serving, whereas the
\lpmsequence{} model contextualizes each transaction within an entity-linked
history. Both variants are pretrained with self-supervised targets that respect
the information available at the relevant decision point.

\subsection{Field-Aware Transaction Representation}
\label{sec:model-field-encoder}

Let a transaction record be a fixed-schema tuple
$\mathbf{x}=(x_c)_{c=1}^{C}$, where $c$ identifies a field and $C$ is the
number of fields in the schema.
We distinguish numerical fields $x_c\in\mathbb{R}$, categorical fields
$x_c\in\{1,\ldots,V_c\}$ with vocabulary size $V_c$, and text fields
$x_c=(x_{c,1},\ldots,x_{c,L_c})$ whose tokens belong to a shared text
vocabulary. Dates, timestamps, postal codes, and descriptions can be
represented through one or more of these types. A field-aware encoder maps every field into a common $D$-dimensional space,
\begin{equation}
	E(\mathbf{x})=\big[e_1(x_1),\ldots,e_C(x_C)\big]^{\mathsf T}
	\in\mathbb{R}^{C\times D}.
	\label{eq:field-encoder}
\end{equation}
Each $e_c$ combines a type-specific value representation with a learned field
identifier. The identifier is retained even when the field value is masked,
allowing the model to distinguish, for example, a masked merchant identifier
from a masked response code. Semantically related fields may share the
parameters of their value encoder. Small field-specific scale and shift
parameters then preserve the role of each field while allowing statistical
strength to be shared across related inputs, such as merchant and card
countries.

\begin{figure*}[t]
	\centering
	\includegraphics[width=\textwidth]{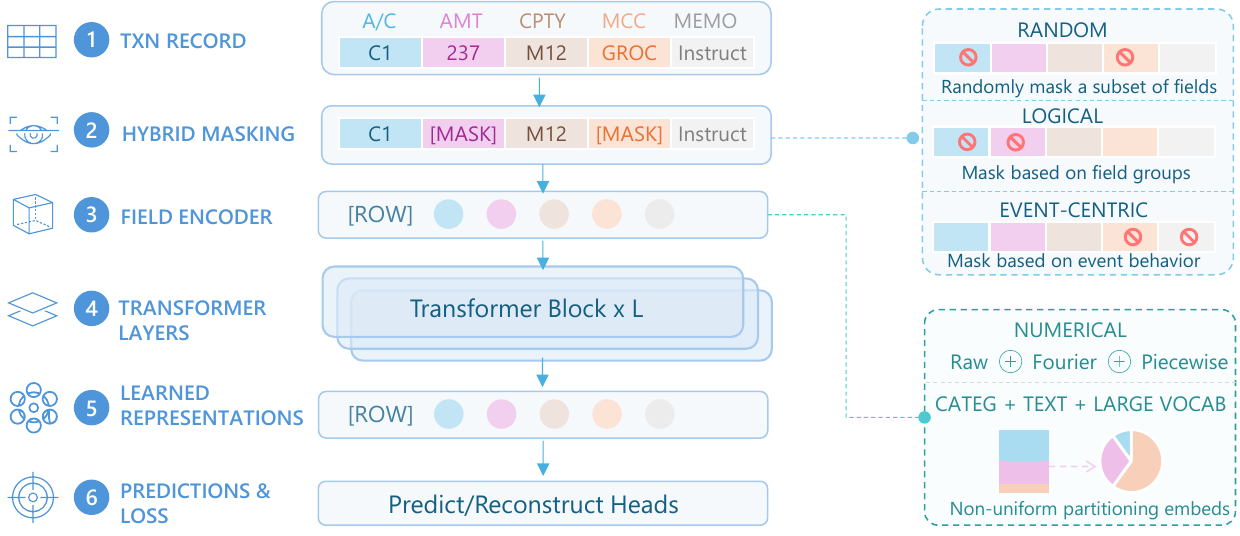}
	\caption{\textbf{LPM \lpmindependent{} model architecture.} During pretraining, hybrid masking is applied to transaction records, after which the field-aware encoder and bidirectional transformer layers produce field representations and a transaction-level representation. Field-specific heads reconstruct masked targets.}
	\label{fig:lpm-independent-architecture}
\end{figure*}

\paragraph{Numerical fields.}
A single linear projection of a scalar does not explicitly encode nonlinear
patterns that matter in payments, including sharp local thresholds and
periodic behavior. We therefore construct a richer representation from raw,
Fourier, and piecewise-linear branches after independently scaling each
numerical field to $[0,1]$ using its empirical quantiles. 
For numerical field $c$, the three branches are concatenated and
projected to the common width as
$e_c^{\mathrm{num}}(x_c)=P_c(r_c(x_c)\oplus f_c(x_c)\oplus
p_c(x_c))$, where $P_c$ and $r_c$ are learned
projections and $\oplus$ denotes concatenation. The Fourier branch is
\begin{equation}
	f_c(x_c)=A_c
	\left[x_c,\{\sin(\omega_{c,j}x_c),
	\cos(\omega_{c,j}x_c)\}_{j=1}^{F}\right],
	\label{eq:numerical-encoder}
\end{equation}
where the frequencies
$\omega_{c,j}$ and Fourier projection matrix $A_c$ are learned. The piecewise
branch uses $K_{\mathrm{pw}}$ ordered centers
$a_{c,1}<\cdots<a_{c,K_{\mathrm{pw}}}$. Both the center locations and their
associated embeddings $\mathbf{v}_{c,k}$ are learned parameters; the learned
locations permit non-uniform spacing and hence variable-width interpolation
intervals. If $x_c$ lies between adjacent centers
$a_{c,k}$ and $a_{c,k+1}$, for $k=1,\ldots,K_{\mathrm{pw}}-1$, the branch
interpolates between their embeddings:
\begin{equation}
	p_c(x_c)=(1-\alpha_c)\mathbf{v}_{c,k}
	+\alpha_c\mathbf{v}_{c,k+1},
	\qquad
	\alpha_c=\frac{x_c-a_{c,k}}{a_{c,k+1}-a_{c,k}}.
	\label{eq:piecewise-encoder}
\end{equation}
The subscript $c$ allows every numerical field to learn its own encoding
parameters.
The Fourier branch provides multi-resolution global variation, while the
piecewise branch provides local resolution around learned breakpoints.

\paragraph{Categorical and text fields.}
Moderate-cardinality categorical fields use learned lookup tables. Text fields
may use token embeddings learned during pretraining or representations from a
frozen pretrained text encoder, depending on the application. The resulting
field representation is mapped to the same width $D$ before being combined
with its field identifier. Distinct learned special tokens represent
out-of-vocabulary, missing, and not-applicable values. This
preserves the difference between a previously unseen value, an absent value,
and a field that is not defined for the transaction type or lifecycle stage.

\paragraph{High-cardinality identifiers.}
Identifiers such as cards, merchants, and accounts can have vocabularies large
enough that a conventional table with $V_cD$ parameters dominates the model.
LPM primarily addresses this problem with a non-uniform partitioned embedding.
The vocabulary is divided into $B_c$ bins, using frequency or an
application-defined allocation rule, and bin $b$ receives width $D_{c,b}$ for
its $V_{c,b}$ values. The embedding storage is therefore
\begin{equation}
	\left|\theta_c^{\mathrm{part}}\right|
	=\sum_{b=1}^{B_c}V_{c,b}D_{c,b},
	\qquad
	\sum_{b=1}^{B_c}V_{c,b}=V_c,
	\label{eq:partitioned-embedding}
\end{equation}
rather than $V_cD$. Larger representations can be assigned to frequent or
operationally important values, while the long tail receives less capacity;
all bins ultimately contribute to the common field space in
Equation~\eqref{eq:field-encoder}. Low-rank factorization and hash-based
embeddings provide alternative operating points. A factorized table uses a
$V_c\times K$ lookup followed by a $K\times D$ projection, requiring
$V_cK+KD$ parameters for $K\ll D$, while hashing removes vocabulary-sized
storage at the cost of collisions. We use partitioning as the principal method
because it preserves distinct representations while allocating capacity
non-uniformly.

\subsection{Independent Transaction Modeling}
\label{sec:model-independent}

The \lpmindependent{} variant models all fields of a transaction jointly but
processes each transaction record independently of other raw records. Historical context,
when used, enters as aggregate fields in $\mathbf{x}$. Inference therefore uses
a fixed-size target record rather than retrieving raw transaction history, and
its cost is independent of history length. \autoref{fig:lpm-independent-architecture}
summarizes the architecture and expands its masking and field-encoding components.

We prepend a learned \texttt{[ROW]} token to the field representations and
apply $L$ transformer blocks with bidirectional self-attention and feed-forward
layers:
\begin{equation}
	H^{(0)}=\left[e_{\mathrm{row}};E(\widetilde{\mathbf{x}})\right],
	\qquad
	H^{(\ell)}=\operatorname{Transformer}_{\ell}
	\left(H^{(\ell-1)}\right),
	\quad \ell=1,\ldots,L,
	\label{eq:independent-forward}
\end{equation}
where $\widetilde{\mathbf{x}}$ is a masked transaction and
$H^{(L)}\in\mathbb{R}^{(C+1)\times D}$. A masked field retains its field
identifier but replaces the value representation with a learned mask
representation. Consequently, each field token can attend to the observed
transaction context, while the final \texttt{[ROW]} token provides a summary
of the complete transaction object.

\paragraph{Hybrid masking.}
A payment record typically combines entity metadata, core event attributes,
aggregates summarizing prior activity, and lifecycle-specific fields. Uniformly
masking individual fields provides broad self-supervision but does not fully
reflect this structure. We therefore sample masks from a mixture of three
strategies. Random masks reconstruct individual fields from the remaining
transaction. Logical span masks remove semantically related groups, such as all
card-related fields, and require their reconstruction from the rest of the
record. Lifecycle-aware masks impose the information
boundary associated with a decision point. For example, post-authorization
outcomes are withheld when the observed context represents the information
available before an authorization decision. The sampling probability may be
biased toward selected fields or groups, and a masking curriculum may vary the
mixture or difficulty of masks over pretraining.

\paragraph{Pretraining objective.}
Let $\mathcal{M}\subseteq\{1,\ldots,C\}$ be the sampled set of masked fields,
$q_c$ the prediction head for field $c$, and $\ell_c$ its prediction loss. An
optional bottleneck $b_c:\mathbb{R}^{D}\rightarrow\mathbb{R}^{d}$, $d\ll D$,
feeds a parallel head $q_c^{\mathrm{bot}}$ with loss
$\ell_c^{\mathrm{bot}}$. We
use cross-entropy for categorical fields, mean-squared error for numerical
fields, and focal loss~\citep{lin2017focal} for highly imbalanced categorical
targets. The \lpmindependent{} pretraining objective is
\begin{equation}
	\mathcal{L}_{\mathrm{ind}}
	=\mathbb{E}_{\mathbf{x},\mathcal{M}}
	\left[
		\frac{\sum_{c\in\mathcal{M}}w_c\!\left[
		\ell_c\!\left(q_c(H_c^{(L)}),x_c\right)
		+\lambda_{\mathrm{bot}}\ell_c^{\mathrm{bot}}\!\left(
		q_c^{\mathrm{bot}}(b_c(H_c^{(L)})),x_c\right)\right]}
		{\sum_{c\in\mathcal{M}}w_c}
	\right],
	\label{eq:independent-loss}
\end{equation}
where $w_c$ optionally adjusts the importance of a field and
$\lambda_{\mathrm{bot}}\geq 0$ controls the auxiliary bottleneck objective;
setting $\lambda_{\mathrm{bot}}=0$ disables it. Only masked values contribute
to either loss. Field-specific heads allow each prediction to use the
appropriate output type, vocabulary size, and loss function. For categorical
fields, prediction-head weights can optionally be tied to the corresponding
input embeddings, avoiding a second parameter matrix and substantially
reducing overhead for large vocabularies.

\paragraph{Output representations.}
The model can expose field-specific predictions, such as classification
logits, as well as the contextual $D$-dimensional field representations that
feed the prediction heads. The final \texttt{[ROW]} state provides a
$D$-dimensional representation of the transaction as a whole. Collecting the
bottleneck outputs produces a compact representation for a column $c$ in
$\mathbb{R}^d$. The auxiliary term in
Equation~\eqref{eq:independent-loss} trains these outputs against the same
masked targets, encouraging them to retain predictive information. They can
then be used as augmented features in downstream models, including tree-based
models for which full-width embeddings may be impractical.

\begin{figure*}[t]
	\centering
	\includegraphics[width=\textwidth]{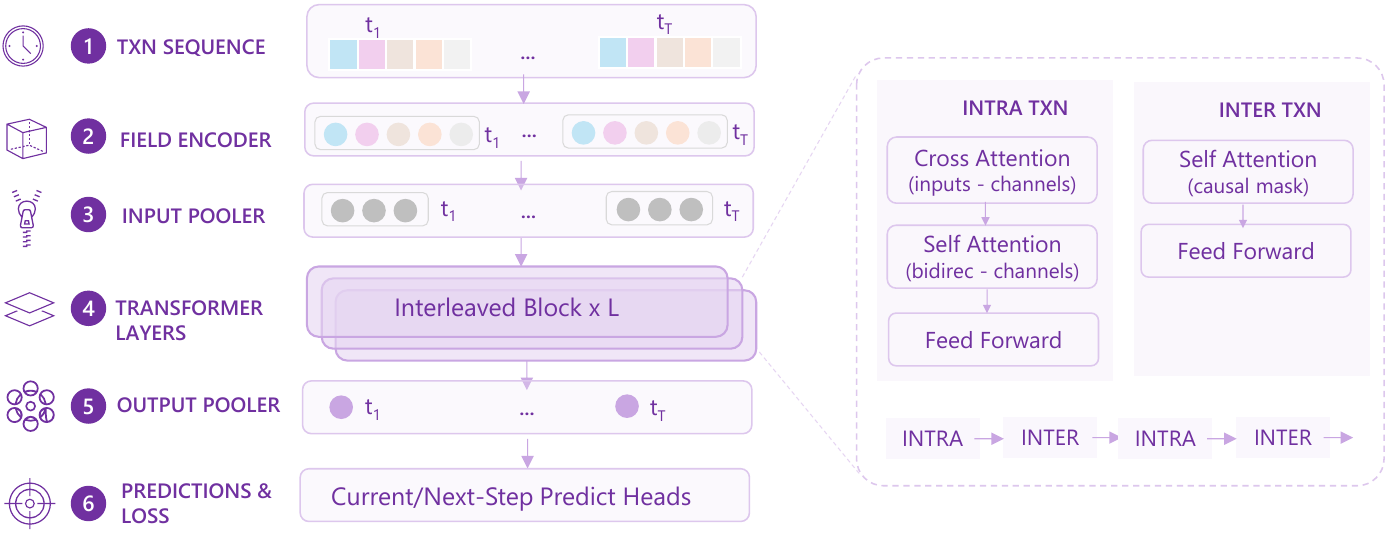}
	\caption{\textbf{LPM \lpmsequence{} model architecture.} An entity-linked transaction history is encoded and pooled into compact latent channels, then processed by interleaved intra-transaction and causal inter-transaction blocks. The former combines cross-attention to the original field representations with self-attention; the latter applies temporal self-attention. Per-step pooling and prediction follow.}
	\label{fig:lpm-sequence-architecture}
\end{figure*}

\subsection{Transaction Sequence Modeling}
\label{sec:model-sequence}

The \lpmsequence{} variant is intended for tasks where aggregate history features do
not capture enough temporal structure. For an entity such as a card, merchant,
or account, let $(\mathbf{x}_1,\ldots,\mathbf{x}_T)$ be a chronologically
ordered transaction history, and let $x_{t,c}$ denote field $c$ of transaction
$\mathbf{x}_t$. Applying the shared field encoder independently at each step
gives $\mathbf{E}=[E(\mathbf{x}_1),\ldots,E(\mathbf{x}_T)]
\in\mathbb{R}^{T\times C\times D}$. Carrying every field through temporal
attention is costly, so an attention pooler compresses each transaction into
$M$ latent channels, where $M\ll C$. \autoref{fig:lpm-sequence-architecture}
summarizes this flow and expands the interleaved transformer blocks.

The sequence network interleaves intra-transaction and inter-transaction
blocks. An intra-transaction block uses bidirectional attention among the $M$
latent channels at a fixed time step and cross-attends to the original $C$
field embeddings. An inter-transaction block applies causal attention through
time independently for each latent channel. Unlike a staged design that
completes all intra-transaction processing before temporal modeling,
interleaving allows temporally updated latents to be repeatedly refined using
within-transaction structure and the original field representations. A compact
representation of this flow is
\begin{align}
	Z^{(0)}
		&=\operatorname{Pool}_{\mathrm{in}}(\mathbf{E}),
	&\overline{Z}^{(\ell)}
		&=\operatorname{CrossIntra}_{\ell}
		  \left(Z^{(\ell-1)},\mathbf{E}\right), \\
	Z^{(\ell)}
		&=\operatorname{CausalInter}_{\ell}
		  \left(\overline{Z}^{(\ell)}\right),
	&\mathbf{H}
		&=\operatorname{Pool}_{\mathrm{out}}
		  \left(Z^{(L)}\right)
		  =\left\{h_{t,0},(h_{t,k})_{k\in\mathcal{K}}\right\}_{t=1}^{T}.
	\label{eq:sequence-forward}
\end{align}
Here, $Z^{(\ell)},\overline{Z}^{(\ell)}\in
\mathbb{R}^{T\times M\times D}$ and
$\mathbf{H}\in\mathbb{R}^{T\times(|\mathcal{K}|+1)\times D}$ for a set
$\mathcal{K}$ of prediction tasks. At each time step, the output pooler
produces a generic representation $h_{t,0}$ and a task-specific representation
$h_{t,k}$ for every $k\in\mathcal{K}$. This avoids requiring all tasks to use a
single pooled representation while retaining a generic per-step output for
reuse.
$\operatorname{CrossIntra}_{\ell}$ operates independently at each time step:
the latent channels attend bidirectionally to one another and use the original
field embeddings as the keys and values of cross-attention.
$\operatorname{CausalInter}_{\ell}$ instead operates through time independently
for each latent channel. This allows the compressed latents to recover detailed
field information without carrying all $C$ tokens through every temporal
operation.
The attention-score complexity of an inter-transaction block is
$O(MT^2)$; choosing $M\ll C$ reduces this cost relative to maintaining one
temporal channel for every field. Causal attention ensures that each pooled
representation at time $t$ depends only on transactions at times $1$ through
$t$.

\paragraph{Dense temporal supervision.}
The \lpmsequence{} model supports current-step and next-event supervision through a
single set $\mathcal{K}$ of application-defined label fields. For each label
$k$, an offset $\delta_k\in\{0,1\}$ specifies whether its head predicts the
field at the current transaction ($\delta_k=0$) or the next transaction
($\delta_k=1$). Thus, a current approval outcome and the next merchant category
can be trained within the same objective. Given a label-specific head $q_k$
and weight $w_k$, let $\rho_{t,k}\geq 0$ denote the temporal weight assigned to
step $t$ for task $k$. The dense objective is
\begin{equation}
	\mathcal{L}_{\mathrm{seq}}
	=\mathbb{E}_{\mathbf{X}}
	\left[
	  \frac{1}{\sum_{k\in\mathcal{K}}w_k}
	  \sum_{k\in\mathcal{K}}
	  \frac{w_k}{\sum_{u=1}^{T-\delta_k}\rho_{u,k}}
	  \sum_{t=1}^{T-\delta_k}
	  \rho_{t,k}\,
	  \ell_k\!\left(q_k(h_{t,k}),x_{t+\delta_k,k}\right)
	\right],
	\label{eq:sequence-loss}
\end{equation}
where $\mathbf{X}=(\mathbf{x}_1,\ldots,\mathbf{x}_T)$ denotes a sampled
transaction history. Uniform weights recover an average over all valid time steps. For selected
tasks, larger weights can instead be assigned to the last few eligible steps,
emphasizing recent outcomes without discarding supervision elsewhere in the
history. The task losses are then combined using $w_k$. A current-step label is
withheld from its own input, while $\delta_k=1$ explicitly shifts a next-event
target by one transaction. Together with causal inter-transaction attention,
these constraints prevent future transactions or later lifecycle outcomes from
leaking into $h_{t,k}$. Because supervision is available at every eligible time
step, a history yields many training signals rather than a single sequence-level
target.

\begin{lpmTakeaway}
The additional context in the \lpmsequence{} model can improve
tasks driven by changes in entity behavior, but deployment requires retrieving,
ordering, and processing a transaction history. The choice between variants is
therefore operational as well as statistical: the \lpmindependent{} model favors predictable
low-latency serving, while the \lpmsequence{} model favors richer temporal context.
\end{lpmTakeaway}

%% file: sec_adap.tex
\section{Downstream Adaptation}
\label{sec:adaptation}
\subsection{Task-Specific Prediction}
\label{sec:model-adaptation}
LPM supports several levels of adaptation. If a downstream target is one of the pretrained fields, the corresponding reconstruction head can be reused under the appropriate availability mask. For a new target, a lightweight task head can be trained on the \texttt{[ROW]} or per-step representation while the backbone is frozen or fine-tuned. When more
capacity is needed, low-rank adaptation modules ~\citep{hu2021lora} can update selected transformer
projections without modifying the full parameter set. For feature-based
downstream models, the \lpmindependent{} model can export the supervised
bottleneck representations introduced in Equation~\eqref{eq:independent-loss},
providing compact field-level features without requiring the full-width
outputs.

\subsection{Knowledge Distillation}
LPM offers optional distillation pathways that transfer predictive information
from a high-capacity teacher to a student chosen for its serving characteristics
~\citep{hinton2015distilling}. For a target
$y$, teacher output $p_T$, and student output $p_S$, we use the general
objective
\begin{equation}
	\mathcal{L}_{\mathrm{distill}}
	=\lambda_{\mathrm{sup}}\mathcal{L}_{\mathrm{sup}}(y,p_S)
	+\lambda_{\mathrm{soft}}\mathcal{D}_{\tau}(p_T,p_S)
	+\lambda_{\mathrm{rank}}\mathcal{L}_{\mathrm{rank}}(p_T,p_S),
	\label{eq:distillation-loss}
\end{equation}
where $\mathcal{L}_{\mathrm{sup}}$ preserves supervision from observed labels,
$\mathcal{D}_{\tau}$ is a student-compatible discrepancy between softened
teacher and student outputs, and the optional ranking term preserves relative
ordering among examples. This objective provides a common template for all
pathways below; individual terms, their coefficients, and the sampling
distribution are selected for the task and student architecture. In
particular, training may emphasize informative examples near the teacher's
decision boundary. Specific student configurations and
distillation settings are evaluated in
Section~\ref{sec:experiments-deployment}.

\paragraph{Within-architecture compression.}
Because differently sized \lpmindependent{} models use compatible field
encoders and prediction interfaces, a compact model can be trained directly
from a higher-capacity teacher. This pathway reduces memory and computation
while preserving the same input schema and serving contract.

\paragraph{Distillation to tree-based models.}
When deployment favors established tree-based pipelines, teacher predictions
can supervise a gradient-boosted decision tree student alongside observed
labels. Soft targets communicate information beyond the observed outcome, while
confidence-aware weighting or label-based correction can limit propagation of
uncertain or incorrect teacher predictions. This pathway trades some
representational capacity for CPU-efficient inference and compatibility with
existing feature-based systems.

%% file: sec_experiments.tex
\section{Experiments}
\label{sec:experiments}

Our experiments assess three aspects of reusable payment intelligence: utility
across multiple tasks, applicability of the LPM approach in
distinct payment domains, and reuse of existing representations for a target
not observed during pretraining. Additional studies
examine representation structure, the effects of data and model scale,
distillation, and the value of sequence context.

\begin{figure*}[t]
	\centering
	\includegraphics[width=\textwidth]{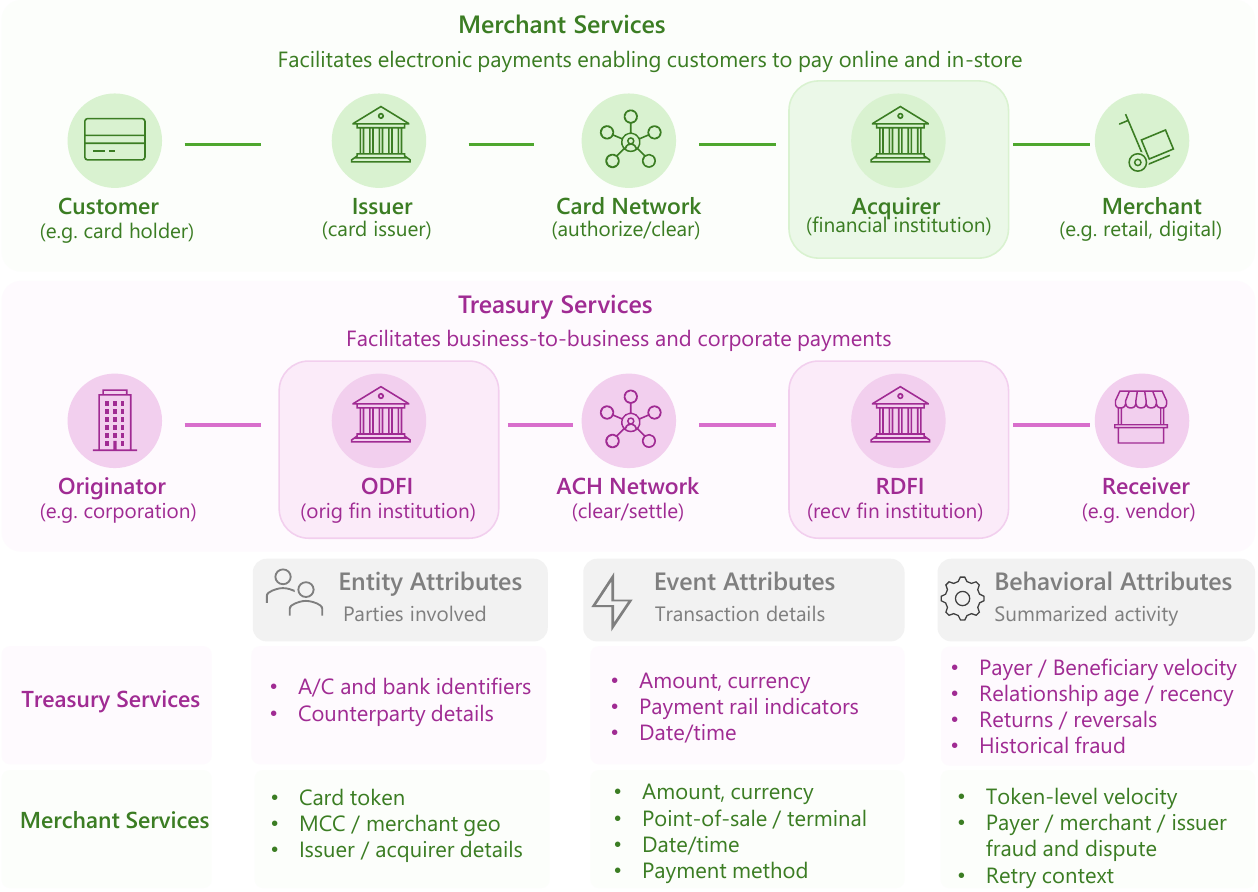}
	\caption{\textbf{Payment domains and representative transaction features.}
	The top and bottom panels schematically identify the principal participants
	in the Merchant Services and Treasury Services payment ecosystems,
	respectively. Highlighted components identify
	JPMorgan Chase's roles as the acquirer, originating and receiving depository financial institution
	(ODFI/RDFI). The accompanying examples show selected event, entity, and behavioral attributes represented in the corresponding transaction records; they are illustrative rather than exhaustive.}
	\label{fig:lpm-features}
\end{figure*}

\subsection{Experimental Setup}
\label{sec:experiments-setup}

\paragraph{Domains and tasks.}
We evaluate LPM in two payment domains, illustrated in
\autoref{fig:lpm-features}. In \emph{Merchant Services}, we consider card transactions
for which JPMorgan Chase acts as the acquirer, spanning card-present and card-not-present
payments across credit, debit, and prepaid rails. We consider three downstream
task families. The primary task, Fraud Detection, estimates whether a transaction will be
identified as fraudulent. Network Response predicts the response returned by
the payment-processing network for a transaction. We evaluate this task at two
label resolutions. The fine-grained variant predicts approximately 100 native
responses, such as approval, insufficient funds, or expiration-date mismatch.
This label space is highly imbalanced, with observations concentrated in a
small number of classes. We also evaluate a coarse-grained variant, for which a
separate model is trained to predict fewer than 10 predefined response
categories. Retry Response concerns outcomes following a decline: whether a
particular reattempt will be approved and whether the transaction will
eventually be approved across subsequent attempts.

In \emph{Treasury Services}, we consider Automated Clearing House (ACH)
payments and wire transfers. We evaluate fraud prevention separately for ACH and wire
payments, which have distinct transaction populations and operational review
budgets.
Together, the experiments span seven prediction settings: five in Merchant
Services and two in Treasury Services.

\paragraph{Transaction schema and features.}
As depicted in \autoref{fig:lpm-features}, each transaction is represented
using three classes of variables. \emph{Event attributes} describe the payment
event itself, including amount, currency, date and time components, and
payment-rail indicators. \emph{Entity attributes} identify and characterize
the parties to the transaction: sender and receiver accounts in Treasury
Services, and the card and merchant in Merchant Services. Representative fields
include card tokens, merchant category codes (MCCs), bank identification
numbers (BINs), member entities, and issuer or acquirer identifiers.
\emph{Behavioral attributes} are engineered aggregates summarizing prior
activity associated with these entities. They include counts, total and
average amounts, and distinct-entity
measures at progressively finer granularities---for example, sender, receiver,
sender--receiver, and sender--receiver--bank combinations---over windows of 7
days, 30 days, 365 days, and 3 years. These aggregates are computed offline
using information available before the transaction's prediction time and
attached to the corresponding record. Both variants learn from these enriched
transaction records, while the \lpmsequence{} variant additionally processes
entity-linked histories. In total, each transaction contains
approximately 150 input features in Merchant Services and 300 in Treasury
Services.

\paragraph{Pretraining corpus construction.}
Production payment volumes exceed the corpus sizes that can be used
practically for pretraining. Uniform random sampling would overrepresent highly
active entities and common transaction patterns while providing limited
exposure to rare but operationally important activity. We therefore construct
the pretraining corpus using entity-aware stratified sampling. The sampler
maintains coverage across activity levels for core entities, including
merchants and cards in Merchant Services and accounts in Treasury Services,
and preserves representation of transaction patterns relevant to the
downstream tasks.

\paragraph{Temporal splits and information controls.}
For each reported experiment, the training data span five months. Evaluation
uses a subsequent two-month out-of-time period separated from the training
period by a gap of at least one month. Inputs exclude features, outcomes, and
lifecycle information unavailable at prediction time. For each
evaluation transaction, the \lpmsequence{} model may additionally use
entity-linked transactions from the preceding six months, restricted to events
occurring before that time.

\paragraph{Model configurations and task prediction.}
We instantiate the \lpmindependent{} and \lpmsequence{} architectures described
in Section~\ref{sec:model}. We use both domains for the main task evaluation,
whereas controlled scaling and adaptation studies focus on Merchant Services,
where matched runs are available to isolate the effects of pretraining data,
model capacity, and adaptation method. Within each architecture, models are
organized into Small, Medium, and Large capacity tiers. These labels describe relative
capacity within an architecture and do not imply matched parameter counts
across architectures or payment domains. The main Merchant Services task
evaluation uses Medium \lpmindependent{} and \lpmsequence{} models pretrained
at approximately the 400B-token scale. The Treasury Services evaluation uses
a Medium \lpmindependent{} model trained on a separately constructed,
domain-specific corpus. The main task evaluations use the corresponding prediction heads learned
during pretraining, without downstream backbone adaptation. Separately, we
study adaptation of the \lpmindependent{} model to a target not observed during
pretraining in Section~\ref{sec:experiments-adaptation}.

\paragraph{Pretraining details.}
Tokenization is performed offline and distributed across workers on multiple
CPU machines. The resulting tokens are stored as sharded datasets, from which
distributed training workers draw examples according to the training sampler.
Models are trained on a multi-node H100 GPU cluster using data-parallel
training.

We optimize both architectures with AdamW, a weight decay of $0.01$, and a
cosine learning-rate schedule with a linear warmup over the first 10\% of
training. The warmup was important for stable optimization. Learning rates
were adjusted as the number of training nodes, and hence the effective global
batch size, increased. The \lpmindependent{} model used global batches in the
hundreds, whereas the
\lpmsequence{} model used smaller batches because each example contains a
transaction history. In the latter, transaction-local operations can process
the $BT$ transaction instances in parallel; only the causal temporal modules
must preserve the $B\times T$ organization. Assigning the temporal modules a
slightly lower learning rate than the remaining components provided a small
optimization benefit.

We evaluated both FP32 training and BF16 automatic mixed precision. FP32
produced better final quality, with a larger gap for the \lpmsequence{} model.
For high-cardinality identifiers, we used up to four non-uniform embedding
partitions with different representation capacities, as described in
Section~\ref{sec:model-field-encoder}. For the \lpmindependent{} objective,
losses for selected task-relevant fields were modestly upweighted through
$w_c$ in Equation~\eqref{eq:independent-loss}. For the \lpmsequence{} model,
the number of latent channels $M$ was set to approximately 20\% of the fields
available in the corresponding schema. We also tested a masking curriculum
that progressively increased the frequency of lifecycle-aware masks, but it
did not yield material gains over a fixed masking mixture.

\paragraph{Evaluation metrics and baselines.}
Evaluation follows the operational decision setting for each task. Each task
is compared with a strong task-specific baseline;
for operational reasons, baseline identities and absolute metric values are not
disclosed.

\emph{Merchant Services.} Results are reported as relative improvement over
the corresponding baseline on the same evaluation population. We use average
precision (AP) uplift for Fraud Detection and Retry Response. For fraud, AP is
well suited to the rare positive class and avoids selecting a single operating
threshold. For the Network Response multiclass tasks, we use macro-averaged F1 as the
primary metric. It computes F1 independently for each response class and
weights all classes equally, preventing frequent responses from dominating the
aggregate score and making performance on rare responses visible despite the
highly imbalanced label distribution.

\emph{Treasury Services.} Fraud prevention on ACH and wire payments is
evaluated under a fixed
alert-review capacity. We therefore report volume-based and value-based recall
at the operating budget. Volume-based recall measures the
fraction of fraudulent transactions captured, whereas value-based recall
measures the fraction of fraudulent transaction value captured, reflecting
both operational feasibility and business impact.

\subsection{Task Evaluation Results}
\label{sec:experiments-task-evaluation}

The Merchant Services results in \autoref{tab:task-evaluation-results} and the
Treasury Services results in \autoref{tab:treasury-task-evaluation-results}
report the best evaluated checkpoint for each architecture--target pair and
fixed-budget target, respectively. Each table follows the task-specific
evaluation protocol defined above. Results are reported separately because the
domains have different prediction objectives, operating constraints, and
metrics.

\begin{table*}[t]
	\centering
	\small
	\renewcommand{\arraystretch}{1.18}
	\setlength{\tabcolsep}{12pt}
	\begin{tabular}{@{}llrr@{}}
		\toprule
		\rowcolor{LPMTable}
		Task & Metric & \lpmindependent{} uplift (\%) $\uparrow$
			& \lpmsequence{} uplift (\%) $\uparrow$ \\
		\midrule
		Fraud Detection                    & AP       & 62.7 & 85.5 \\
		Network Response (fine-grained)    & F1 Macro & 80.0 & 89.5 \\
		Network Response (coarse-grained)  & F1 Macro & 4.5  & 8.4 \\
		Retry Response                     & AP       & 22.8  & 22.8 \\
		\bottomrule
	\end{tabular}
	\caption{Merchant Services task evaluation results. Values report
	relative uplift in the row-specific metric under the evaluation setup
	described in Section~\ref{sec:experiments-setup}.}
	\label{tab:task-evaluation-results}
\end{table*}

\begin{table*}[t]
	\centering
	\small
	\renewcommand{\arraystretch}{1.18}
	\setlength{\tabcolsep}{12pt}
	\begin{tabular}{@{}lrr@{}}
		\toprule
		\rowcolor{LPMTable}
		Task & Recall Volume uplift (\%) $\uparrow$
			& Recall Value uplift (\%) $\uparrow$ \\
		\midrule
		ACH fraud  & 27.4 & 63.5 \\
		Wire fraud & 5.6  & 0.0 \\
		\bottomrule
	\end{tabular}
	\caption{Treasury Services fraud-prevention results at fixed
	alert-review budgets. Values report relative uplift over
	the corresponding baseline under the same evaluation population
	and fixed review capacity.}
	\label{tab:treasury-task-evaluation-results}
\end{table*}

In Merchant Services, both LPM variants improve over the corresponding
task-specific baselines across all four settings. The largest gains occur in
Fraud Detection and fine-grained Network Response. The latter is particularly
notable because it requires distinguishing about 100 highly
imbalanced response classes. The \lpmindependent{} model improves relative
macro-F1 by 80\%, and the \lpmsequence{} model adds nearly 10
percentage points of uplift. Because macro-F1 weights each class equally, the
result is not dominated by the most frequent responses. On Fraud Detection,
the sequence-aware configuration similarly provides 23
percentage points of additional AP uplift over the independent model. It also provides
higher uplift on coarse-grained Network Response, while the two variants
perform similarly on Retry Response. The magnitude of the sequence-aware
advantage is therefore task-dependent.

On ACH payments, LPM delivers substantial gains at a fixed daily alert-review budget, increasing volume-based recall by about 27\% and value-based recall by about 64\% relative to the operational baseline. This improvement surfaces more fraudulent transactions and substantially more fraudulent value without increasing review capacity. On wire payments, LPM produces a more modest improvement, with about 6\% uplift in volume-based recall.  The smaller gains may reflect the lower representation of wire payments in the pretraining corpus and gaps in signals available for some transactions. These findings motivate wire-specific data enrichment and model adaptation as priorities for further work.

\begin{lpmTakeaway}
LPM's pretrained prediction heads improve fraud detection, Network Response,
and Retry Response over task-specific baselines in Merchant Services. A
separately trained Treasury Services model also improves fraud detection,
extending the approach beyond card payments. The benefits are task-dependent:
sequence context adds substantial value for Merchant Services fraud detection, while
at a fixed ACH review budget, relative gains in fraudulent value captured
exceed those in fraudulent transaction count.
\end{lpmTakeaway}

\subsection{Adaptation to a New Payment Task}
\label{sec:experiments-adaptation}

The preceding results evaluate prediction targets learned during pretraining.
We next test whether an existing LPM representation can support an additional
target: whether a declined transaction that is subsequently retried will
eventually be approved. This target was not observed during pretraining. The
study uses a Medium \lpmindependent{} model pretrained on
approximately 85B tokens as the common backbone. This target differs from Retry Response in
\autoref{tab:task-evaluation-results}, which predicts the outcome of a
particular reattempt. All adaptation methods use the same downstream training
set and out-of-time evaluation population as the task-specific baseline.

\autoref{tab:task-adaptation-results} compares three increasingly expressive
ways to adapt the \lpmindependent{} model. The linear probe trains logistic
regression on fixed \texttt{[ROW]} embeddings, which can be computed offline
and reused across tasks. The attention-pooling variant keeps the backbone
frozen but learns to combine different field representations for the new target. The
LoRA variant updates low-rank adapters together with a classifier on the
\texttt{[ROW]} representation. AP is reported as the relative change from the
task-specific baseline.

\begin{table*}[t]
	\centering
	\small
	\renewcommand{\arraystretch}{1.18}
	\setlength{\tabcolsep}{6pt}
	\begin{tabular*}{\textwidth}{@{}lclrl@{}}
		\toprule
		\rowcolor{LPMTable}
		Adaptation method & Backbone & Parameters
			& Relative $\Delta$AP $\uparrow$ & Practical value \\
		\midrule
		Task-specific baseline              & --     & --   & Reference & Dedicated solution \\
		Linear probe on \texttt{[ROW]}       & Frozen & $<1$K & $-2.1\%$ & Reusable representations \\
		Attention pool over field embeds & Frozen & $\sim$110K & $-1.3\%$ & Learned field aggregation \\
		LoRA with \texttt{[ROW]} head        & LoRA   & $\sim$670K & $+0.5\%$ & Efficient backbone adaptation \\
		\bottomrule
	\end{tabular*}
	\caption{Adaptation to eventual retry approval, a downstream prediction target
	unavailable during pretraining. Relative AP change is computed against the task-specific
	baseline; all methods use the same downstream training set and out-of-time
	evaluation population. Trainable parameters exclude the frozen LPM backbone.}
	\label{tab:task-adaptation-results}
\end{table*}

With fewer than 1,000 trainable parameters, the linear probe comes within 2.1\% of the
task-specific baseline. It therefore provides an immediate head start for a
new task through \texttt{[ROW]} embeddings that can be computed offline and
reused, without adapting the LPM backbone. Learning a task-specific attention
pool over field embeddings narrows the remaining gap while leaving the
backbone unchanged. LoRA reaches parity with the task-specific model and
slightly exceeds it in this evaluation, using approximately 670K trainable
parameters.

\begin{lpmTakeaway}
For eventual retry approval, a target not observed during pretraining, frozen
LPM representations approach a dedicated task-specific baseline and compact
low-rank adaptation achieves comparable performance. The result demonstrates
reuse of an existing backbone beyond its original pretraining objectives.
\end{lpmTakeaway}

\subsection{Representation Space Analysis}
\label{sec:experiments-qual}

\begin{figure*}[t]
	\centering
	\includegraphics[width=\textwidth]{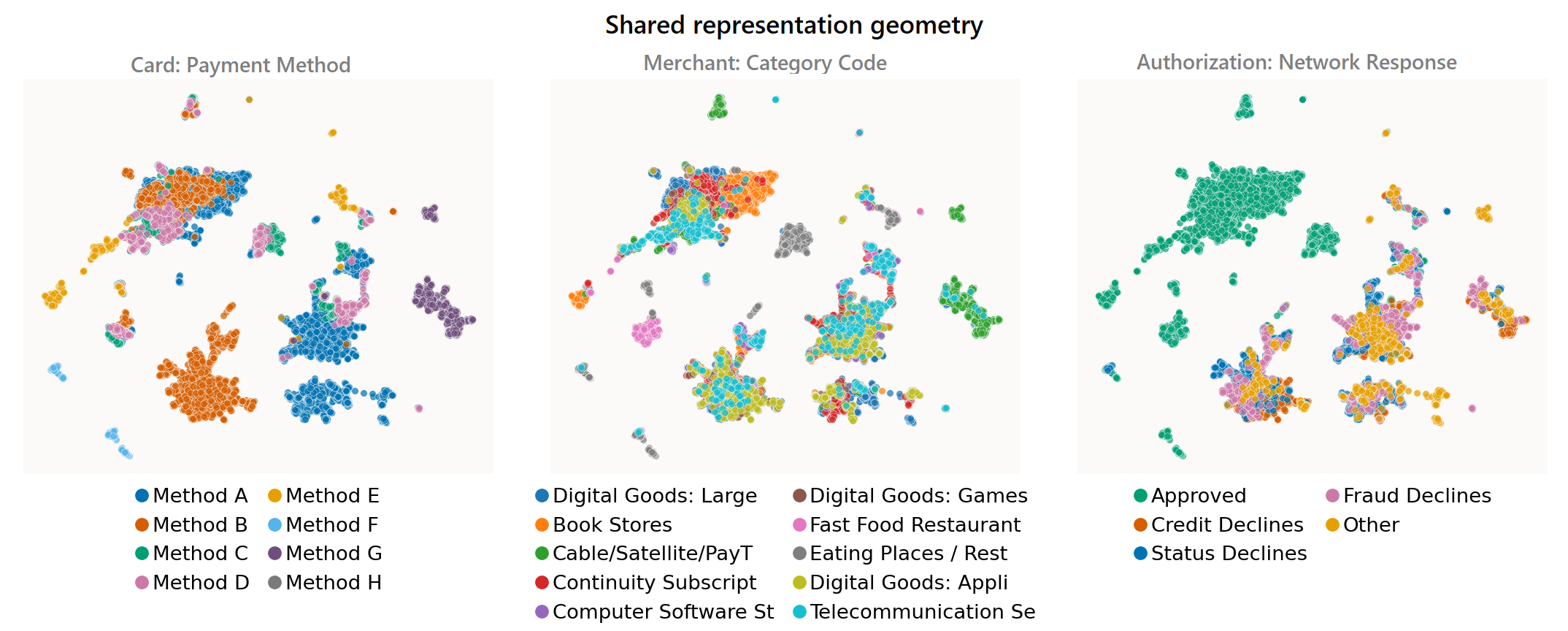}
	\caption{\textbf{Multiple payment semantics in a shared transaction
	representation.} Each panel shows the same UMAP projection of 30,000
	\texttt{[ROW]} embeddings, with colors defined by payment method, merchant category, and network response.
	Payment-method names are anonymized. The recurring local structure across
	views shows that embedding neighborhoods reflect multiple facets of a
	transaction.}
	\label{fig:lpm-representation-umap}
\end{figure*}

\paragraph{Embedding visualization.}
We examine the final \texttt{[ROW]} representations produced by the
\lpmindependent{} model for 30,000 out-of-time Merchant Services transactions.
The representations are extracted under the authorization-time information
boundary: Network Response and later lifecycle fields are masked, whereas
payment method and MCC are observed inputs. We fit one two-dimensional UMAP
projection and reuse the same transactions and coordinates in all three panels
of \autoref{fig:lpm-representation-umap}, changing only the attribute used to
color the points. Payment methods are anonymized for operational reasons.

The projection exhibits partial, overlapping organization rather than
artificially clean class boundaries. Network Response shows the clearest
large-scale structure, with approval and decline outcomes concentrated in
different regions. Payment methods form several localized regions while also
overlapping in a large shared region, and selected merchant categories show
weaker but visible local concentrations. This mixture is expected because a
\texttt{[ROW]} representation summarizes the complete transaction rather than
encoding a single taxonomy, and because a two-dimensional projection cannot
preserve all neighborhoods from the original representation space.

\paragraph{Neighborhood analysis.}
To quantify local organization without relying on the UMAP projection, we
evaluate $k$-nearest-neighbor classification directly in the original
768-dimensional representation space. We randomly sample 30,000 test
transactions and divide them into a 20,000-transaction reference set and a
10,000-transaction query set. For each query, we identify its $k$ nearest
reference embeddings by cosine distance and predict the attribute by majority
vote. \autoref{tab:representation-knn} reports macro-F1 over the displayed
categories. Values outside a displayed category set remain eligible reference
neighbors as an ignored category but are not included among scored queries.
The permuted-label null retains the query labels, embedding geometry, class
frequencies, and neighbor assignments, but randomly reassigns labels among the
reference embeddings; the reported null is the mean over ten permutations at
$k=30$.

\begin{table*}[t]
	\centering
	\small
	\renewcommand{\arraystretch}{1.18}
	\setlength{\tabcolsep}{10pt}
	\begin{tabular}{@{}lrrr@{\hspace{18pt}}r@{}}
		\toprule
		\rowcolor{LPMTable}
		Attribute & \multicolumn{3}{c}{Observed macro-F1 $\uparrow$}
			& Permuted-label null \\
		\cmidrule(lr){2-4}\cmidrule(l){5-5}
		& $k=10$ & $k=30$ & $k=100$ & $k=30$ \\
		\midrule
		Payment method   & 0.919 & \textbf{0.889} & 0.773 & 0.098 \\
		Merchant category & 0.660 & \textbf{0.604} & 0.488 & 0.079 \\
		Network Response & 0.800 & \textbf{0.793} & 0.762 & 0.163 \\
		\bottomrule
	\end{tabular}
	\caption{Neighborhood structure of Merchant Services \texttt{[ROW]}
	representations. Observed columns report cosine $k$-NN macro-F1 in the
	original representation space. The final column reports the mean macro-F1
	after permuting labels across reference embeddings while holding query labels
	and neighborhoods fixed.}
	\label{tab:representation-knn}
\end{table*}

At $k=30$, observed macro-F1 exceeds the permuted-label null by 0.791 for
payment method, 0.525 for merchant category, and 0.630 for Network Response.
The first two results show that the shared representation retains substantial
information about observed card and merchant attributes. More notably, the
Network Response label is masked during representation extraction, so its
strong and comparatively stable neighborhood score indicates that the
surrounding transaction context organizes the representation according to an
unobserved authorization outcome. Together, the projection and neighborhood
analysis provide complementary evidence that a single transaction
representation captures several payment-relevant semantics.

\subsection{Scaling Behavior}
\label{sec:experiments-scaling}

Both scaling studies focus on Merchant Services and use prediction heads
learned during pretraining, without downstream backbone adaptation.

\paragraph{Pretraining data scale.}

\begin{figure*}[t]
	\centering
	\includegraphics[width=\textwidth]{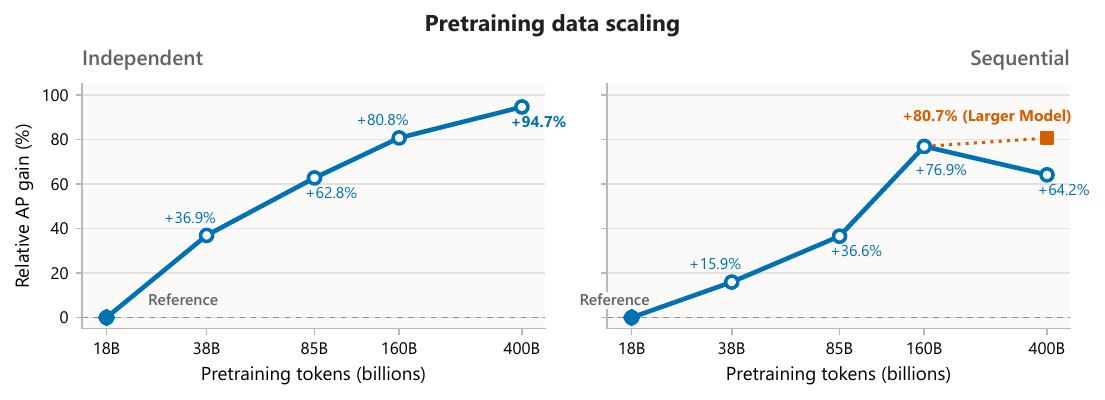}
	\caption{Pretraining data scaling on Merchant Services fraud detection.
	The left and right panels show \lpmindependent{} and \lpmsequence{} models,
	respectively. Each series reports the relative change in AP compared with
	the 18B-token Medium configuration of the same architecture. The orange
	result highlights additional gains from a larger \lpmsequence{} model at
	400B tokens.}
	\label{fig:lpm-data-scaling}
\end{figure*}

We study data scaling on Merchant Services fraud detection using Medium
\lpmindependent{} and \lpmsequence{} models and pretraining budgets of 18B,
38B, 85B, 160B, and 400B tokens. Within each Medium-model series, the model
configuration and evaluation setting are fixed while only the pretraining
corpus size varies.
\autoref{fig:lpm-data-scaling} reports the relative AP gain at each scale,
normalized to the 18B-token Medium configuration of the same architecture. Separate
panels present the \lpmindependent{} and \lpmsequence{} trends without implying
that their configurations are parameter matched.

The \lpmindependent{} model improves steadily with additional pretraining data,
reaching a 95\% AP gain over its 18B-token reference at 400B tokens. The
\lpmsequence{} model also benefits from additional data, although its gains
are non-monotonic at fixed capacity. In particular, at 400B tokens, a
larger-capacity model achieves an AP gain of about 81\%, surpassing the Medium
model's best result and highlighting the interaction between data scale and
model capacity.

\paragraph{Model capacity.}

\begin{figure*}[t]
	\centering
	\includegraphics[width=\textwidth]{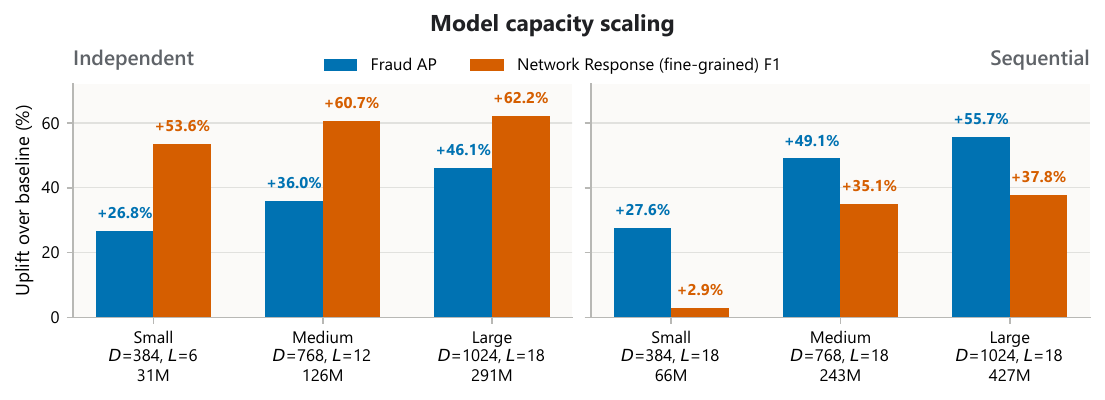}
	\caption{Model capacity scaling at approximately 85B pretraining tokens.
	The left panel shows \lpmindependent{} model's capacity tiers, which vary both
	representation width and depth; the right panel shows \lpmsequence{} model's
	configurations with depth fixed at 18 layers and varying representation
	width. Bars report relative uplift in Fraud Detection AP and fine-grained
	Network Response macro-F1 over their
	respective task-specific baselines.}
	\label{fig:lpm-model-scaling}
\end{figure*}

We next vary the available capacity tiers of both architectures while holding
the pretraining corpus at approximately 85B tokens.
\autoref{fig:lpm-model-scaling} reports Fraud
Detection AP and fine-grained Network Response macro-F1 uplift over their
respective task-specific baselines. The capacity paths differ by
architecture: \lpmindependent{} varies both representation width and depth,
whereas \lpmsequence{} holds depth fixed at 18 layers and varies representation
width. The figure reports each configuration's parameter count, and trends are
interpreted within each architecture.

Increasing model capacity improves both tasks within each architecture, with
the Large configuration achieving the strongest result in every comparison.
The largest change occurs for fine-grained Network Response in the
\lpmsequence{} model, where uplift rises from about 3\% for Small to 35\% for
Medium, followed by a more modest gain for Large. This pattern suggests that
the value of additional capacity depends on both the architecture and task.

\subsection{Distillation}
\label{sec:experiments-deployment}

We evaluate knowledge distillation in the Treasury Services fraud setting,
using AP as the evaluation metric and a Medium LPM as the teacher. The teacher is trained on the full
transaction corpus and assigns each training transaction a fraud probability.
These probabilities determine both the examples included in a reduced
distillation training set, or \emph{transfer set}, and the soft targets used
to train the students. All fraud transactions are retained. The selected
non-fraud portion comprises 90\% high-scoring transactions and 10\% transactions
sampled uniformly at random. All student variants use the same transfer set,
so their comparison is not confounded by differences in training-example
selection.

We consider two student architectures. The GBT student is an XGBoost model on
the raw transaction features, with depth 6, learning rate 0.05, subsampling
rate 0.8, minimum child weight 1.0, $L_2$ regularization 1.0, and 2,000
boosting rounds. The Small LPM uses the teacher architecture at reduced
capacity. It receives a one-epoch reconstruction warm start followed by two
epochs of distillation.

Both distilled students instantiate the objective in
Equation~\eqref{eq:distillation-loss}, using cross-entropy to match the teacher's fraud
probability. For the GBT, $\lambda_{\mathrm{sup}}=0.5$ and
$\lambda_{\mathrm{soft}}=0.5$. For the Small LPM,
$\lambda_{\mathrm{sup}}=0.1$ and $\lambda_{\mathrm{soft}}=0.9$, with focal
reconstruction of the fraud field as the supervised term. We additionally
train a hard-label GBT on the same transfer set, without the
teacher-probability term, to isolate the contribution of soft supervision.

\begin{table}[t]
	\centering
	\small
	\renewcommand{\arraystretch}{1.18}
	\setlength{\tabcolsep}{8pt}
	\begin{tabular}{@{}llr@{}}
		\toprule
		\rowcolor{LPMTable}
		Model & Supervision & Relative $\Delta$AP (\%) $\uparrow$ \\
		\midrule
		LPM (Medium) & Teacher    & Reference \\
		LPM (Small)  & Distilled  & $-5.8$ \\
		GBT          & Distilled  & $-16.7$ \\
		GBT          & Hard label & $-24.9$ \\
		\bottomrule
	\end{tabular}
	\caption{Distillation results for Treasury Services fraud detection.
	Values report the relative change in AP from the
	Medium LPM teacher; all student variants use the same transfer set.}
	\label{tab:distillation-results}
\end{table}

As shown in \autoref{tab:distillation-results}, the Small LPM retains the most
teacher performance, finishing about 6\% below the Medium teacher. The
distilled GBT trails the teacher by about 17\%, but improves by roughly 8
percentage points over the hard-label GBT trained on the identical transfer
set. Thus, the teacher's soft probabilities provide useful supervision beyond
transfer-set selection alone. The Small LPM's roughly 11 percentage-point
advantage over the distilled GBT quantifies the trade-off between the two
students. Distillation therefore allows the student architecture to be
selected according to serving constraints: the Small LPM offers greater
fidelity to the teacher, while the GBT supports established tree-based
serving stacks.

\subsection{Ablation Studies}
\label{sec:experiments-ablations}

\paragraph{Sensitivity to sequence length.}
We study how much transaction context is needed by the \lpmsequence{} model by
varying the sequence length $T$ over $1$, $2$, $4$, $8$, $16$, $24$, and $32$
transactions. The ablation covers Merchant Services Fraud Detection and both
resolutions of Network Response. Because the tasks use different metrics, each
score is normalized to the corresponding result at $T=32$; the reference
configuration therefore has a value of 100\% in every panel. 

\begin{figure*}[t]
	\centering
	\includegraphics[width=\textwidth]{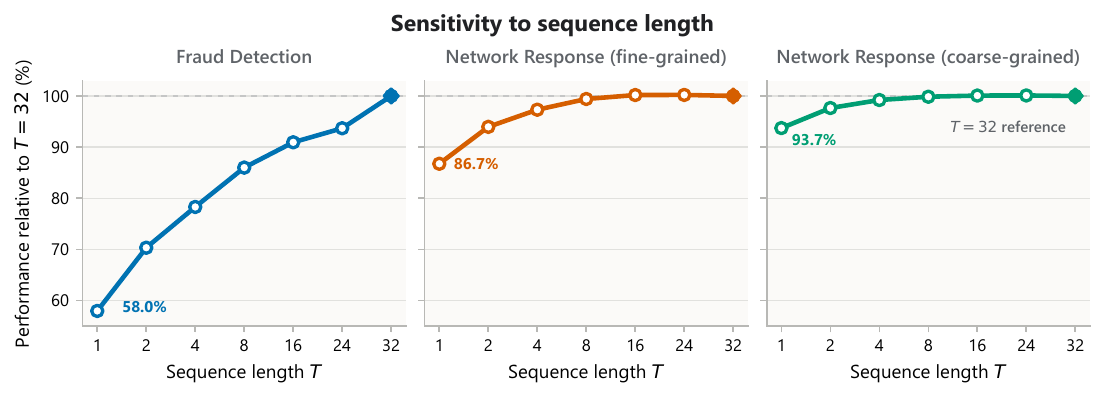}
	\caption{Sensitivity of the \lpmsequence{} model to sequence length on
	Merchant Services tasks. The panels report Fraud Detection AP and
	fine- and coarse-grained Network Response macro-F1, respectively. Within
	each task, performance is expressed relative to the score obtained with
	$T=32$ transactions.}
	\label{fig:lpm-seqlen-ablation}
\end{figure*}

As shown in \autoref{fig:lpm-seqlen-ablation}, the value of additional context
depends strongly on the task. Fraud Detection benefits throughout the tested
range: a single-transaction sequence retains 58\% of the $T=32$ AP, while
$T=8$ and $T=24$ retain 86\% and about 94\%, respectively. Network Response
saturates much earlier. At $T=8$, both Network Response resolutions retain more than 99\% of their
respective $T=32$ macro-F1, with only small
fluctuations thereafter. The fine-grained task is more sensitive at very short
lengths, retaining about 87\% at $T=1$, compared with about 94\% for the coarse-grained
task. These results
indicate that longer histories are most useful for fraud patterns, whereas
response prediction obtains nearly all of its measured benefit from a short
recent context.

%% file: sec_deploy.tex
\section{Operational Deployment}
\label{sec:deployment}

\begin{figure*}[t]
	\centering
	\includegraphics[width=\textwidth]{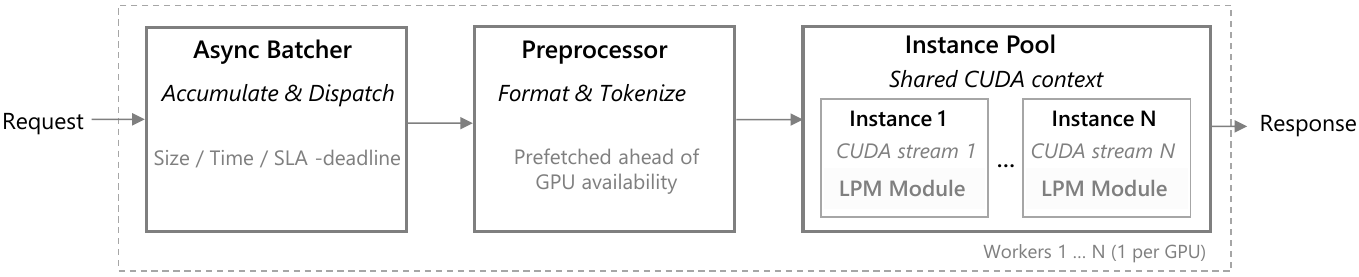}
	\caption{Overview of inference flow through batching, preprocessing, and the model instances.}
	\label{fig:lpm-deploy}
\end{figure*}

\paragraph{Serving architecture.}
The production inference service (\autoref{fig:lpm-deploy}) is designed to
maximize GPU utilization, a
primary cost driver at scale. Multiple in-process model instances share a CUDA
context and use dedicated CUDA streams, allowing concurrent calls to overlap
kernel execution without repeated context-switch overhead. An adaptive batcher
dispatches requests according to both batch size and queueing time, balancing
throughput against latency. Additional optimizations include ONNX Runtime graph
fusion (including fused attention, SkipLayerNorm, and FastGelu), an optional
TensorRT compilation path, and GIL-free CPU tokenization in dedicated
subprocesses with prefetching and low-overhead inter-process payloads.

\paragraph{Deployment and model lifecycle.}
The service runs on Amazon EKS with the NVIDIA GPU Operator providing the CUDA
runtime at the node level, reducing hardware-specific container management.
Kubernetes readiness probes admit traffic only after all workers have loaded
and warmed their models, protecting rolling deployments and autoscaling events.
Model versions and artifacts are managed through MLflow and staged on a shared
FSx for Lustre volume at pod startup. This separation of model lifecycle from
serving supports champion--challenger and shadow deployments, in which a
candidate receives a configurable share of live or shadow traffic before
promotion.

\begin{lpmTakeaway}
The system serves millions of transactions per day with p99
latency below $60\,\mathrm{ms}$ and accommodates bursts of up to 500
transactions per second without breaching its service-level objective. At this
workload, stream-level GPU sharing and adaptive batching keep effective
inference cost below US\$0.0001 per transaction. These results indicate that
transformer inference can meet the latency, throughput, resilience, and cost
requirements of payment-scale deployment.
\end{lpmTakeaway}

%% file: sec_findings.tex
\section{Discussion, Limitations and Future Work}
\label{sec:discussion}

\paragraph{Evidence for reusable payment intelligence.}
The evaluations provide complementary evidence for a reusable payments
intelligence layer. Merchant Services results demonstrate
utility across multiple pretrained prediction objectives, from rare-event
detection to highly imbalanced multiclass prediction, while Treasury Services
results establish applicability of the approach in a distinct payment domain.
More directly, adaptation to a target not observed during pretraining shows
that an existing backbone can support an additional task through compact
components. Production deployment adds operational evidence that predictive
signals can be delivered under payment-scale latency, throughput, and cost
constraints. These findings link reusable learning with downstream use and
practical serving.

\paragraph{Cross-payments learning.}
Merchant and Treasury Services currently use separately trained domain models,
reflecting their distinct schemas, transaction populations, and operational
objectives. Building on this demonstrated applicability, a common
cross-payments representation is the next empirical test of how far the
intelligence layer can extend. Comparison with domain-specific models would
assess whether shared learning improves transfer across payment types while
preserving performance within each domain.

\paragraph{Task-dependent operating points.}
Temporal context produces its largest incremental gains for Merchant Services
fraud detection and fine-grained Network Response, while other tasks show
smaller gains or saturate with short histories. The
\lpmsequence{} model is most compelling when these gains justify the cost of
retrieving and processing richer histories. Elsewhere, the
\lpmindependent{} model offers a strong alternative with a simpler serving path
and inference cost independent of raw history length. The scaling studies
similarly favor matching data volume and model capacity to the architecture
and task.

\paragraph{Data modeling as model design.}
Field identities and lifecycle-aware masks express constraints specific to
payment records, while entity-aware sampling makes large transaction volumes
tractable without concentrating the corpus on the most active entities and
most common events. LPM's learned representations incorporate engineered
aggregates. Their contribution relative to event and entity fields and raw
histories remains to be isolated through controlled ablations.

\begin{lpmTakeaway}[before skip=4pt, after skip=4pt]
At payment scale, schema coverage, information availability, and corpus
composition are modeling choices rather than preprocessing details.
\end{lpmTakeaway}

\paragraph{Design boundaries.}
A fixed schema preserves field semantics and supports efficient computation,
but extension to new fields or independently defined schemas may require
additional pretraining; typed encoders and schema-extension mechanisms could
reduce this cost. Neither variant fully removes the need to summarize long
histories: the \lpmindependent{} model uses aggregate features, and the
\lpmsequence{} model uses a bounded context. Learned memory offers one
direction for representing longer-term activity. Generative applications,
including synthetic transaction generation, remain outside the present scope.

%% file: sec_conclusion.tex
\section{Conclusion}
\label{sec:conclusion}
We introduced the Large Payments Model (LPM), a domain-specific foundation
model that learns representations directly from structured payment
transactions. LPM combines field-aware encodings and lifecycle-aware
self-supervision with two complementary architectures: the
\lpmindependent{} model for efficient per-transaction inference and the
\lpmsequence{} model for incorporating richer transaction histories. The results demonstrate 
utility across multiple prediction tasks in two payment domains, adaptation of an existing
representation to a target not observed during pretraining, and
operational feasibility under real-time constraints. Although the present evaluation 
is limited to selected domains and tasks, these results
show that large-scale pretraining tailored to the structure and operational
constraints of payments can provide a reusable payments intelligence layer.